\documentclass[lettersize,journal]{IEEEtran}
\usepackage{amsmath,amsfonts}
\usepackage{array}
\usepackage[caption=false,font=normalsize,labelfont=sf,textfont=sf]{subfig}
\usepackage{textcomp}
\usepackage{stfloats}
\usepackage{url}
\usepackage{verbatim}
\usepackage{graphicx}
\usepackage{cite}
\usepackage[hidelinks]{hyperref}
\hypersetup{colorlinks=true,citecolor=blue,linkcolor=blue,urlcolor=blue}
\usepackage{enumitem}
\usepackage{amssymb}
\usepackage{balance}

\usepackage{tikz}
\usepackage{cleveref}
\crefformat{section}{\S#2#1#3}
\crefformat{subsection}{\S#2#1#3}
\crefformat{subsubsection}{\S#2#1#3}
\DeclareMathOperator*{\argmax}{arg\,max}

\newcommand{\circlelabel}[1]{\tikz[baseline=(char.base)] \node[shape=circle,draw,inner sep=0.6pt] (char) {#1};}
\newcommand{\bluecite}[1]{\unskip \textcolor{blue}{\cite{#1}}}

\usepackage[linesnumbered,ruled,vlined, noend]{algorithm2e}

\usepackage{bm}
\usepackage{dsfont}

\begin{document}

\title{QuickGrasp: Responsive Video-Language Querying Service via Accelerated Tokenization and Edge-Augmented Inference}

\author{Miao Zhang,~\IEEEmembership{Member,~IEEE,} Ruixiao Zhang, Jianxin Shi, Hengzhi Wang,~\IEEEmembership{Member,~IEEE,} Hao Fang, and Jiangchuan Liu,~\IEEEmembership{Fellow,~IEEE}

\thanks{Miao Zhang, Hao Fang, and Jiangchuan Liu are with School of Computing Science, Simon Fraser University, Canada. Ruixiao Zhang is with Computer Science Department, University of Illinois Urbana-Champaign, United States. Jianxin Shi is with College of  Cryptology and Cyber Science, Nankai University, China. Hengzhi Wang is with College of Computer Science and Software Engineering, Shenzhen University, China.
}
}



\maketitle

\begin{abstract}
Video-language models (VLMs) are reshaping video querying services, bringing unified solutions to complex perception and reasoning tasks. However, deploying large VLMs in real-world systems remains challenging due to their high resource demands, and remote-based deployment often results in unacceptable response delays. Although small, locally deployable VLMs offer faster responses, they unavoidably fall short in accuracy. To reconcile this trade-off, we propose QuickGrasp, a responsive, quality of service (QoS)-aware system that bridges this gap through a local-first architecture with on-demand edge augmentation. Built upon the highly modular architecture of VLMs, QuickGrasp shares the vision representation across model variants to avoid redundant computation. To maximize system-wide efficiency, QuickGrasp introduces three key designs: accelerated video tokenization, query-adaptive edge augmentation, and delay-aware, accuracy-preserving vision token density configuration. We implement a prototype of QuickGrasp and evaluate it across multiple video understanding benchmarks. The results show that QuickGrasp matches the accuracy of large VLMs while achieving up to a 12.8x reduction in response delay. QuickGrasp represents a key advancement toward building responsive video querying services for open-world understanding that fully leverage the capabilities of VLMs.
\end{abstract}


\section{Introduction}

\IEEEPARstart{R}{ecent} breakthroughs in video-language models (VLMs), such as Video-ChatGPT\bluecite{Maaz2024VideoChatGPT}, VideoChat\bluecite{li2023videochat}, and Video-LLaVA\bluecite{lin2024video}, have revolutionized the landscape of intelligent video querying services. This shift moves the paradigm from passive analytics based on predefined pipelines to interactive, on-demand question answering characterized by human-like reasoning. VLMs take as input a user query comprising a video and a textual question, and generate answers that support diverse open-world tasks ranging from attribute perception to complex temporal reasoning. Such capabilities position VLMs as key technical enablers for next-generation video-centric services, including embodied service robots, smart video assistants, and intelligent forensic retrieval\bluecite{duan2022survey, tang2025video}. In these interactive scenarios, a critical metric for evaluating the quality of service (QoS) is \emph{response delay}. Low response delay is essential not only for user convenience but also for the practical viability of these applications. For example, when a user queries a smart video assistant to identify a specific action or locate an event within a video, delayed reasoning severely breaks conversational fluidity and renders the interactive tool impractical. Studies show that a response delay exceeding 1 second begins to interrupt the user's flow of thought, and a delay longer than 10 seconds often leads to service abandonment\bluecite{response-delay}.

\begin{figure}[!t]
    \centering
    \includegraphics[width=0.93\linewidth]{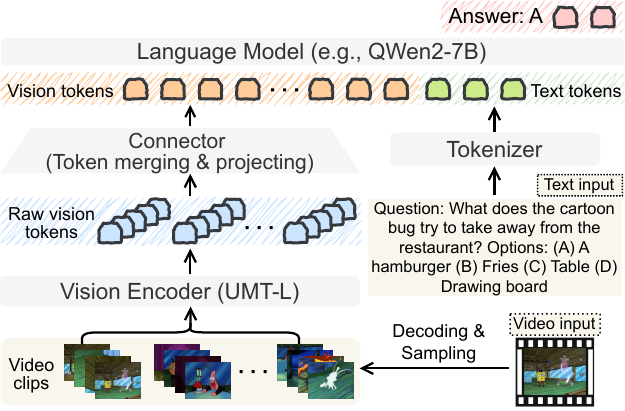}
    \caption{Architecture of the VideoChat-Flash models. The LM can be replaced by smaller or larger ones.}
    \label{fig:videochat-flash}
\end{figure}

However, achieving low response delay is challenging. VLMs are inherently resource-intensive due to the dual challenges of massive model parameters and high-dimensional video inputs. As shown in Fig.~\ref{fig:videochat-flash}, a VLM typically comprises three modular components: a vision encoder (often 100-500 million parameters), a cross-modal connector, and a language model (LM, often 2-13 billion parameters). The inference process begins with video tokenization, where the raw video is decoded, sampled, and transformed by the vision encoder into spatiotemporal embeddings. This stage can incur substantial delay, driven by the heavy overhead of decoding video streams and the computational burden of extracting features from a massive volume of sampled frames. These vision tokens are then compressed and projected into the LM's embedding space. Finally, the LM processes the projected tokens from all frames simultaneously along with the tokenized textual question. This \emph{all-at-once} processing strategy, combined with the parameter scale of the LM, imposes significant GPU memory requirements and computational delays.

To address these substantial resource requirements, proprietary Model-as-a-Service (MaaS) platforms, such as ChatGPT\bluecite{gpt-4o} and Gemini\bluecite{google-gemini}, predominantly offload the entire query execution to the remote cloud. However, this approach shifts the bottleneck from computation to communication. Transmitting data-intensive user queries over networks to remote servers introduces a severe response delay. For instance, based on our measurements using the video understanding benchmark \emph{Video-MME}\bluecite{fu2025videomme}, network transmission alone incurs an average delay of about 15 seconds under today's typical upload speed of 59 Mbps\bluecite{speedtest}. This overhead is prohibitive for interactive services, critically impairing QoS and user satisfaction.

Deploying local VLM inference services offers a compelling alternative that eliminates network transmission bottlenecks. Nevertheless, this approach faces strict resource provisioning constraints, as client-side service nodes (e.g., consumer-grade GPUs) often lack sufficient memory to host large-scale VLMs. These nodes are practically restricted to VLMs with roughly 2B parameters or fewer, as deploying larger models can trigger immediate out-of-memory failures. Consequently, recent architectural optimizations have focused on developing compact VLMs tailored for these constrained service environments\bluecite{zhu2025comprehensive, marafioti2025smolvlm, Shaker2025MobileVideoGPT}. These efforts have yielded promising results regarding service fidelity, where our measurements reveal that a compact 2B VLM generates answers identical to a larger 7B counterpart in 60\%--76\% of cases. While this indicates that a significant portion of queries can be handled locally, the performance gap on complex queries suggests that the higher reasoning capabilities of larger models remain indispensable. This observation motivates a natural \textit{local-first} service orchestration strategy that prioritizes local execution for simple queries while escalating to a remote large VLM only when necessary.

While promising, a naive local-first service orchestration strategy does not automatically guarantee strict QoS compliance. Translating this architectural potential into tangible end-to-end (E2E) response delay reduction requires overcoming three practical challenges. These challenges span from intrinsic VLM processing bottlenecks to collaborative trade-offs. First, independent of the deployment scheme, the delay bottleneck often originates from the VLM's video tokenization stage, where transforming high-dimensional video data into tokens imposes severe delays before the LM inference begins. Second, within the collaborative framework, determining when to escalate a query is non-trivial, as indiscriminately offloading a query negates the responsiveness benefits of local execution. Third, even when offloading is required, balancing the volume of transmitted data against the need for accuracy creates a complex optimization problem.

To address these challenges, we present \textit{QuickGrasp}, a responsive video-language querying service built upon a local-first architecture with on-demand edge augmentation. Unlike monolithic cloud-based solutions or rigid distributed pipelines that split every query execution, QuickGrasp prioritizes execution on the local device to minimize the response delay. This design treats the proximal edge server purely as an accuracy enhancement facility, ensuring that the system incurs network overhead only when the local model's confidence is insufficient. Our key contributions are summarized as follows:

\begin{itemize}[label=$\diamond$, leftmargin=*]
    \item \textbf{Accelerated Video Tokenization.} User queries incorporate videos that vary widely in duration, ranging from a few seconds to several hours. Our measurements reveal that as the video duration grows, video decoding and frame sampling introduce substantial delays. While often overlooked in conventional delay optimization research, these preprocessing stages can become the dominant contributor to the E2E response delay. QuickGrasp mitigates the bottleneck with a decoding-efficient tokenization design that combines keyframe-aligned sampling with pipelined video-to-token conversion, achieving significant acceleration, particularly for long-duration videos.

    \item \textbf{Query-Adaptive Edge Augmentation.} To reduce response delay and computational costs when edge augmentation is triggered, QuickGrasp introduces a collaborative architecture that shares the vision representation across models deployed locally and at the edge. This design allows the edge model to directly reuse the local video tokenization results, avoiding redundant video-to-token conversion and enabling the system to offload compact vision tokens rather than raw video. Maximizing the utility of this shared architecture requires accurately identifying queries that need edge involvement. Unlike static pre-execution routing strategies that rely solely on textual analysis to predict query difficulty, QuickGrasp leverages the calibrated inference confidence of the local VLM to accurately gauge the necessity for edge augmentation on the fly.

    \item \textbf{QoS-Aware Token Density Configuration.} In the token-based offloading pathway, QuickGrasp dynamically selects the vision token density, defined as the average number of per-frame vision tokens fed into the LM. This density directly trades off offloading delay against LM generation quality. A fixed density is often suboptimal, incurring unnecessary network and computational overheads for simple queries or providing insufficient visual evidence for complex reasoning. QuickGrasp formulates token density selection as a contextual multi-armed bandit (CMAB) problem and adapts the density online to maximize a long-term utility that balances answer quality against response delay.

    \item \textbf{Prototype Implementation and Real-World Evaluation.} We implement a fully functional service prototype of QuickGrasp and deploy it on commercial platforms to validate its real-world feasibility. We evaluate QuickGrasp on three comprehensive video understanding benchmarks comprising thousands of videos, with durations ranging from seconds to hours. We also compare against state-of-the-art service baselines. Results show that QuickGrasp matches the accuracy of remote large-scale VLMs while reducing response delay by up to 12.8$\times$.
    
\end{itemize}

The remainder of this article is organized as follows. \cref{sec:background} provides background and motivates the problem. \cref{sec:system-design} presents the design of QuickGrasp. \cref{sec:prototype} describes the service prototype implementation, and \cref{sec:evaluation} reports the evaluation results. \cref{sec:related-work} reviews related work, and \cref{sec:conclusion} concludes the article.

\section{Background and Motivation}
\label{sec:background}

\subsection{Preliminary: Video-Language Models}

\begin{figure}[!t]
    \centering
    \includegraphics[width=\linewidth]{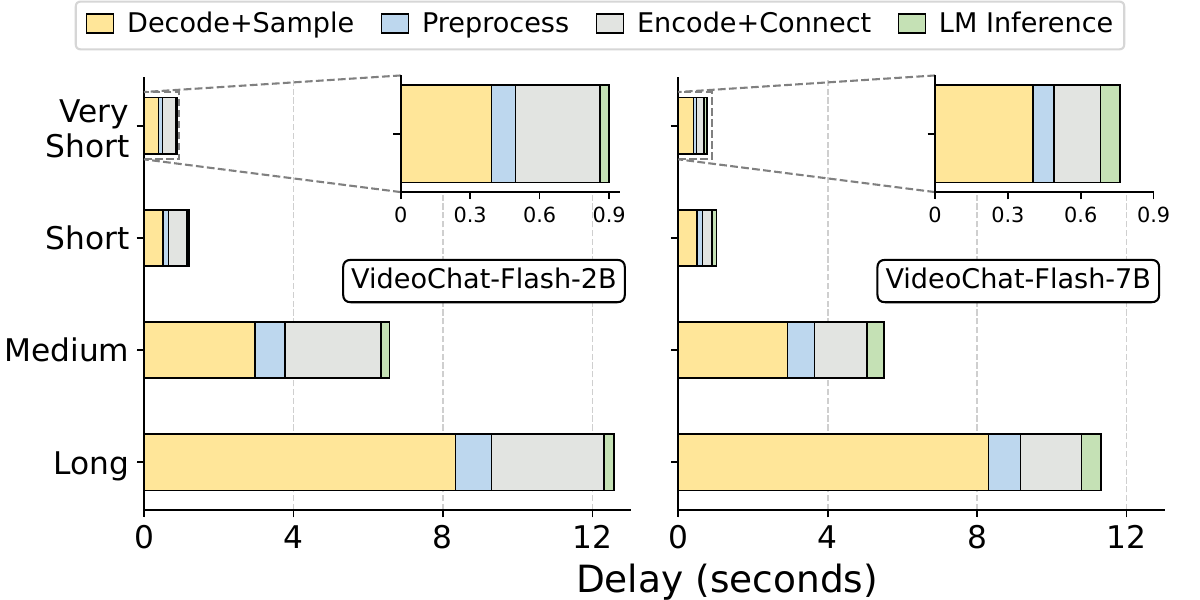}
    \caption{Inference delay breakdown for the 2B and 7B VLM variants.}
    \label{fig:latency_breakdown}
\end{figure}

Although VLMs differ in implementation details, they generally follow an architecture comprising a vision encoder, a lightweight connector, and an LM. The vision encoder is typically a vision Transformer pretrained on massive image-text or video-text pairs, such as CLIP\bluecite{radford2021learning} and UMT\bluecite{li2023unmasked}. The connector can be as simple as a linear layer\bluecite{liu2023visual} or a multilayer perceptron (MLP)\bluecite{liu2024improved}, which projects the vision tokens generated by the vision encoder into the LM's embedding space. Since most LMs have a limited context window, the connector often incorporates a token reduction module, such as Resampler\bluecite{alayrac2022flamingo}, Q‑Former\bluecite{li2023blip2}, spatiotemporal pooling\bluecite{Maaz2024VideoChatGPT}, or token merging\bluecite{li2024videochat}, to reduce the number of vision tokens fed into the LM.

A representative example is VideoChat-Flash\bluecite{li2024videochat}, whose architecture is illustrated in Fig.~\ref{fig:videochat-flash}. During inference, the model employs fixed-rate sampling to extract video frames at regular temporal intervals such as 1 frame per second. These sampled frames are grouped into local clips and processed in batches by the vision encoder, specifically UMT-L\bluecite{li2023unmasked}, to produce spatiotemporal embeddings or vision tokens. The connector then applies token merging\bluecite{bolya2023token} to reduce the vision token count and projects these tokens into the LM's input embedding space. In parallel, the textual question is tokenized by the LM tokenizer. The resulting text tokens are concatenated with the projected vision tokens and subsequently fed into the LM to generate the final answer.

The evaluation of VLMs typically adheres to one of the two paradigms: open-ended or closed-ended question answering (QA). Open-ended QA requires the model to generate free-form textual responses, which are then assessed by human evaluators or large language models (LLMs) such as ChatGPT\bluecite{Maaz2024VideoChatGPT}. Nevertheless, the intrinsic diversity in phrasing and content of open-ended answers introduces ambiguity, making reproducibility and leaderboard comparisons difficult. Hence, most contemporary VLM benchmarks adopt a closed-ended, multiple-choice QA format\bluecite{li2024mvbench, fu2025videomme, zhou2025mlvu}, where the VLM is prompted to generate the correct answer from a predefined set of options (e.g., `A' in Fig.~\ref{fig:videochat-flash}). This format produces standardized outputs and enables more objective, interpretable, and comparable evaluations. Therefore, this study focuses on closed-ended evaluation, reserving open-ended assessments for future research.

\begin{figure}[!t]
    \centering
    \includegraphics[width=\linewidth]{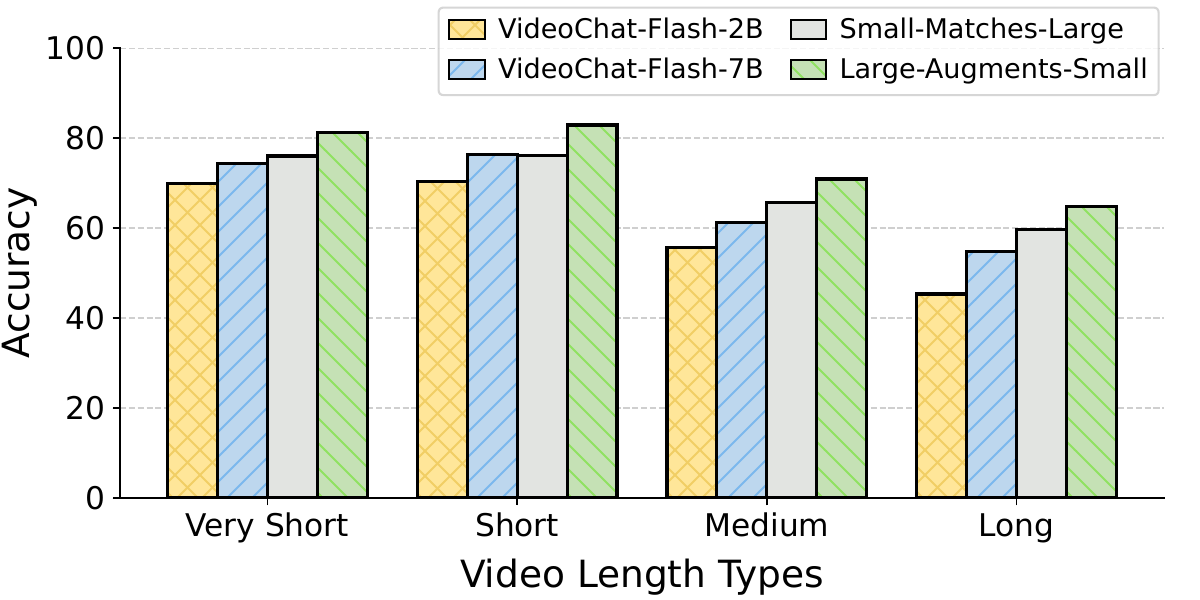}
    \caption{Accuracy comparison on different lengths of videos.}
    \label{fig:model_accuracy_comparison}
\end{figure}

\subsection{Identifying the Delay Bottleneck in VLM Inference}
\label{sec:motivation-latency}
To develop responsive VLM-based video querying services, it is critical to identify the primary delay bottlenecks within the VLM inference pipeline. We adopt VideoChat-Flash as our representative case study, given its strong performance across diverse benchmarks. Specifically, we profile its 2B variant on a local device equipped with a consumer-grade NVIDIA GeForce RTX 4090 GPU (24 GB) and its 7B variant on an edge server equipped with an NVIDIA RTX PRO 6000 GPU (96 GB)\footnote{Detailed hardware specifications are provided in \cref{sec:prototype}. The 7B variant is benchmarked on the edge server because its memory requirements for long-video inference exceed the 24 GB capacity of the local device, leading to out-of-memory errors.}. To ensure reasonable accuracy for spatial reasoning tasks, the input resolution is fixed at 448$\times$448 for both models.

Since VideoChat-Flash uses fixed-rate sampling, the number of processed frames grows linearly with video duration. This linear growth can directly translate into an increased response delay for VLM inference. We therefore examine how this scaling behavior manifests across a wide range of video lengths. Specifically, we evaluate two video understanding benchmarks that cover both short-form and long-form videos. The first is \emph{MVBench}\bluecite{li2024mvbench}, which contains 3,687 videos with durations from 5 to 35 seconds. We refer to this set as \emph{very short} in the following comparisons. The second is Video-MME\bluecite{fu2025videomme}, which consists of 900 videos. We adopt the original duration categories from this benchmark, which include short (under 2 minutes), medium (4 to 15 minutes), and long (30 to 60 minutes).

Fig.~\ref{fig:latency_breakdown} presents the mean inference delay breakdown for the 2B and 7B VLM variants. As video duration increases, the overall inference delay for both variants rises substantially. Under fixed-rate sampling, the number of sampled frames grows linearly with video length. This growth directly increases video tokenization and subsequent processing costs. In particular, the mean delay for long videos is approximately 14$\times$ that of very short videos for the 2B variant and 15$\times$ for the 7B variant. These results highlight the critical need for more efficient frame sampling mechanisms to reduce the response delay.

Furthermore, despite the use of common acceleration techniques such as efficient random frame access and multithreaded decoding, the decoding and sampling stage remains the primary bottleneck. As Fig.~\ref{fig:latency_breakdown} reveals, this preprocessing stage consistently dominates the pipeline and accounts for 42\% to 73\% of the total inference delay across model variants and video durations. Interestingly, this significant overhead is often overlooked in prior VLM inference optimization efforts. Most existing works focus primarily on accelerating vision encoding \bluecite{vasu2025fastvlm} or LM generation \bluecite{tao2025dycoke}. These findings motivate the need to explicitly address the bottlenecks inherent in video decoding and frame sampling.

\subsection{Analyzing the Accuracy Gap Across VLM Variants}
We further examine the accuracy of different model variants across varying video lengths. As shown in Fig.~\ref{fig:model_accuracy_comparison}, both variants exhibit notable accuracy degradation once video duration extends beyond the \emph{short} category. Moreover, the performance gap between the two variants increases for longer videos. This suggests that extended temporal contexts introduce additional challenges that affect the smaller model more significantly. Despite this, the 2B variant remains competitive on very short and short videos. To quantify prediction alignment, we use the 7B model's predictions as a proxy ground truth and evaluate the consistency of the 2B model. The results, represented by the \emph{Small-Matches-Large} bars in Fig.~\ref{fig:model_accuracy_comparison}, reveal that the 2B model produces identical responses to the 7B model in approximately 60\% to 76\% of cases. This high overlap indicates that invoking the large model is unnecessary for many queries. These findings motivate a \emph{local-first} strategy in which the large model is triggered only when the small model is likely to fail.

To characterize the upper bound of such a collaborative strategy, we consider an idealized \emph{Oracle} setting. In this scenario, a routing mechanism perfectly determines whether the 2B variant's output is correct and invokes the 7B variant only when the 2B variant fails. We evaluate the achievable accuracy under this assumption, which serves as a theoretical ceiling for collaborative inference strategies. The resulting accuracy, denoted by the \emph{Large-Augments-Small} bars in Fig.~\ref{fig:model_accuracy_comparison}, exceeds that of the standalone 7B variant. This performance gain suggests a notable complementarity between the two variants. Specifically, the 2B variant occasionally answers queries correctly that the 7B variant misses, which may stem from differences in inductive biases, attention to task-relevant cues, or generalization behavior. Overall, these results indicate that collaborative inference can offer a higher accuracy ceiling than relying solely on the 7B variant.

\begin{figure}[!t]
    \centering
    \includegraphics[width=\linewidth]{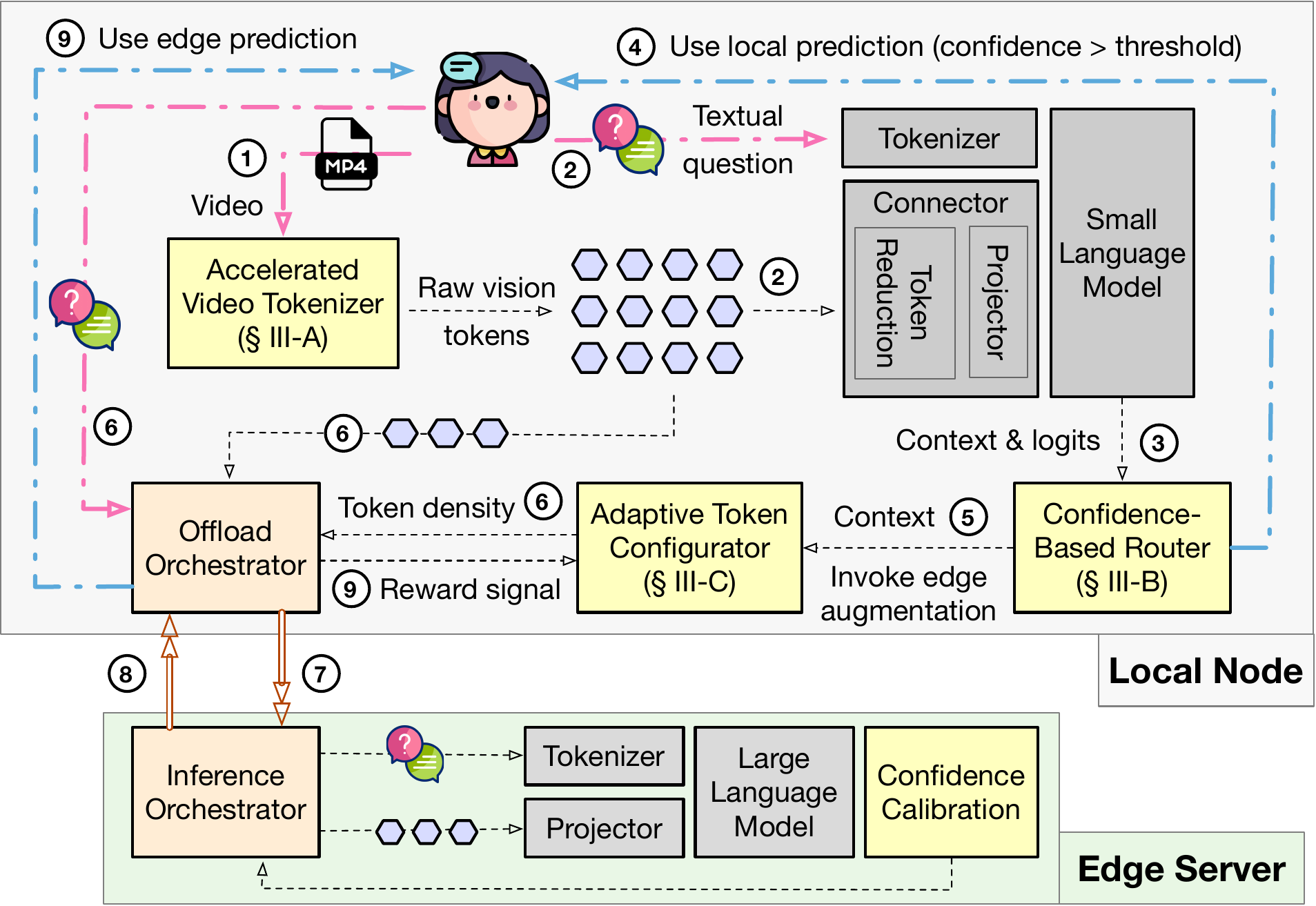}
    \caption{System Overview.}
    \label{fig:system-overview}
\end{figure}

\section{QuickGrasp: System Design}
\label{sec:system-design}

An overview of QuickGrasp is presented in Fig.~\ref{fig:system-overview}. For each input query, the video is first processed by the Accelerated Video Tokenizer (\cref{sec:decode-optimize}) to obtain raw vision tokens (\circlelabel{1} in Fig.~\ref{fig:system-overview}). This component utilizes keyframe-aligned sampling and pipelined video-to-token conversion to eliminate the decoding bottlenecks typically associated with long-duration video inputs. These tokens, along with the textual question, are then passed to the local small VLM for initial inference \circlelabel{2}. To ensure service fidelity, the Confidence-Based Router (\cref{sec:confs-design}) assesses the local prediction by calculating a calibrated confidence score \circlelabel{3}. If the calibrated confidence exceeds a predefined threshold, the local result is accepted as the final answer \circlelabel{4}. 

Otherwise, edge augmentation is triggered, and the local inference context is passed to the Adaptive Token Configurator (\cref{sec:token-configuration}) to determine the offloading scheme \circlelabel{5}. Instead of offloading the original user query, the system leverages shared vision representations to avoid repetitive computation and reduce network transmission delays. Based on the configurator's decision regarding the target token density \circlelabel{6}, the local raw vision tokens are merged into the target count and losslessly compressed by the Offload Orchestrator for transmission \circlelabel{7}. Once the data is received at the edge, it is handled by the Inference Orchestrator and processed by the Large VLM to generate a high-accuracy response. This edge prediction is returned to the local node as the final answer \circlelabel{8}-\circlelabel{9}, while also serving as a reward signal for the Adaptive Token Configurator to facilitate online policy updates via a CMAB framework \circlelabel{9}. The following sections detail the key designs.

\subsection{Accelerated Video Tokenization}
\label{sec:decode-optimize}
\noindent \textbf{Keyframe-Aligned Sampling.} Input videos within user queries can encompass a high volume of frames. For instance, a one-hour video at 30 FPS consists of 108k frames, making it infeasible to ingest every frame in a single VLM forward pass due to memory and context-window limits. \emph{Uniform sampling} is therefore commonly used to reduce the frame count, typically via two primary strategies. \emph{Fixed-count sampling} \bluecite{dai2023instructblip, lin2024video} selects a constant number of frames regardless of duration. While this yields a predictable cost, it becomes overly sparse for long videos and potentially misses key events. In contrast, \emph{fixed-rate sampling} \bluecite{zhang2025llavamini, li2024videochat} samples at regular intervals such as 1 FPS. This method better preserves temporal coverage but incurs substantial decoding and sampling delay as videos grow longer, as discussed in \cref{sec:motivation-latency}. Recent efforts have attempted to further filter uniformly sampled frames using auxiliary models such as CLIP \bluecite{radford21clip} to retain only those semantically relevant to the question \bluecite{zhang2025qframe, tang2025adaptive}. While this may reduce the final sampled frame count, the filtering process itself can introduce substantial computational overhead. Such overhead makes this approach unsuitable for building responsive services.

To overcome these limitations without incurring extra model overhead, the Accelerated Video Tokenizer in QuickGrasp adopts a \emph{keyframe-aligned sampling} strategy that leverages the video's compression structure to reduce decoding delay. Modern codecs (e.g., H.264/HEVC) organize frames into groups of pictures (GOPs), where random access is only efficient at \emph{keyframes} or I-frames. Accessing a non-keyframe requires the decoder to fetch and process all preceding reference frames within the GOP, introducing significant computational and I/O overhead. By aligning sampling indices with these random-access points, we avoid decoding large numbers of reference frames that are not eventually sampled. Moreover, video encoders often insert I-frames at moments of major visual change such as scene cuts or sudden motion. Thus, keyframe-aligned sampling also preserves informative event transitions and serves as a natural, lightweight content filter.

\begin{algorithm}[!t]
\small
\KwData{$\mathit{video\_path}$; $\mathcal{N}_{min}$; $\mathcal{N}_{max}$;}
\KwResult{The indices of frames to be sampled}
$vr \gets Open(\mathit{video\_path})$ \;
Video frame count $\mathcal{N}_{video} \gets len(vr)$ \;
\If{$\mathcal{N}_{video} < \mathcal{N}_{min}$}{
    $sample\_indices \gets range (0, \mathcal{N}_{video})$ \;
}
\Else{
    $key\_indices \gets vr.get\_keyframe\_indices()$ \;
    $\mathcal{N}_{key} \gets len(key\_indices)$ \;
    \If{$\mathcal{N}_{key} > \mathcal{N}_{max}$}{
        $sample\_indices \gets$ Uniformly sample $\mathcal{N}_{max}$ frames from $key\_indices$ \;
    }
    \ElseIf{$\mathcal{N}_{key} < \mathcal{N}_{min}$}{
        $sample\_indices \gets$ Uniformly sample frames from the video at 1 FPS \;    
    }
    \Else{
        $sample\_indices \gets$ $key\_indices$ \;
    }
}
Return $sample\_indices$ \;
\caption{Keyframe-Aligned Sampling Strategy}
\label{alg:keyframe}
\end{algorithm}

The details of our sampling strategy are presented in Algorithm \ref{alg:keyframe}, where $\mathcal{N}_{min}$ and $\mathcal{N}_{max}$ denote the minimum and maximum numbers of frames to be sampled, respectively.\footnote{Throughout this paper, we set $\mathcal{N}_{min}=64$ and $\mathcal{N}_{max}=512$ to align with the default configuration of VideoChat-Flash.} For short videos with fewer than $\mathcal{N}_{min}$ frames, all frames are sampled directly (lines 3-4). For longer videos, the algorithm first retrieves the keyframe indices from the video container metadata without performing actual decoding (line 6). If the number of keyframes exceeds $\mathcal{N}_{max}$, the algorithm uniformly samples $\mathcal{N}_{max}$ keyframes (lines 8-9). If the number of keyframes is below $\mathcal{N}_{min}$, the algorithm falls back to fixed-rate sampling (lines 10-11). In all other cases, all available keyframes are selected (lines 12-13). This algorithm ensures that decoding remains strictly aligned with random-access points whenever possible, while maintaining robust coverage across varying video lengths.

\begin{figure}[!t]
    \centering
    \includegraphics[width=\linewidth]{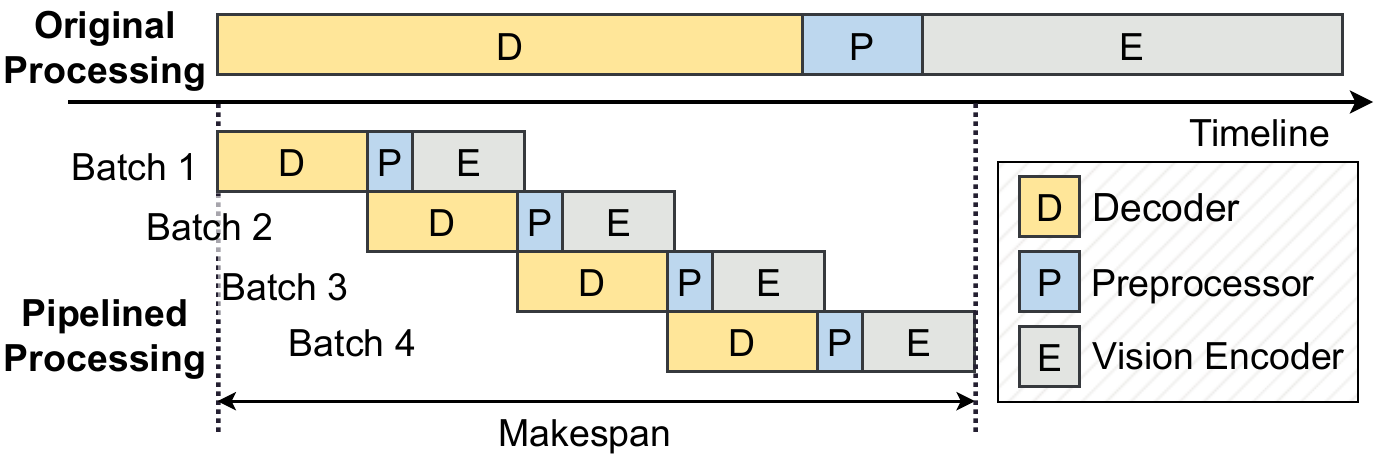}
    \caption{Example demonstrating how pipelined processing mitigates inference delay bottlenecks in VLMs.}
    \label{fig:decode-accelerate}
\end{figure}

\noindent \textbf{Pipelined Video Tokenization.} Vision encoders in VLMs are typically implemented either as image encoders (e.g., CLIP), which encode each video frame independently, or as video encoders (e.g., UMT-L), which group frames into local clips first and then encode each clip separately. In either case, the input units (frames or clips) are inherently independent, eliminating the need for sequential dependency during inference. Building on this observation, the Accelerated Video Tokenizer further integrates a \emph{pipelined processing} strategy that overlaps decoding, preprocessing, and vision encoding to shorten the overall makespan. Instead of waiting for the entire video to be decoded, the strategy operates as a continuous pipeline. The decoder incrementally yields frame batches that are immediately ingested by the preprocessor, whose output is then consumed by the vision encoder. As illustrated in Fig.~\ref{fig:decode-accelerate}, this strategy effectively hides the delay of overlapping stages, significantly decreasing the overall processing time.

\begin{figure}[!t]
    \centering
    \includegraphics[width=\linewidth]{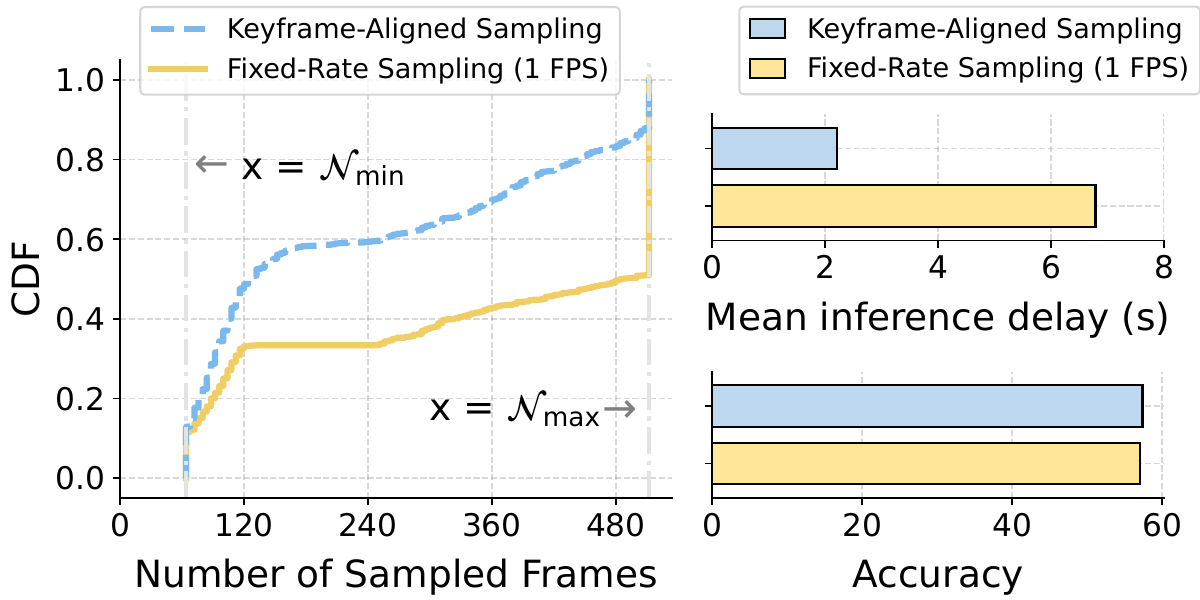}
    \caption{Comparison of different decoding and sampling strategies (Model: VideoChat-Flash-2B; Platform: local node).}
    \label{fig:sample_strategies}
\end{figure}

\noindent \textbf{Pilot Experiment.} Since keyframes are relatively sparse in videos, keyframe-aligned sampling may substantially decrease the sampled frame count, raising valid concerns about potential accuracy degradation. To assess this impact, we conducted a pilot experiment on the Video-MME benchmark, applying the proposed keyframe-aligned sampling and pipelined processing techniques. As shown in Fig.~\ref{fig:sample_strategies}, the CDFs reveal a substantial reduction in the number of sampled frames: the median drops from 486 under fixed-rate sampling to 124 with keyframe-aligned sampling. This reduction, combined with pipelined processing, decreases the mean inference delay by about 67$\%$. Interestingly, the mean accuracy improves slightly from 57.1 to 57.4 despite the reduced frame count. This suggests that keyframes capture the most informative content, and the exclusion of redundant frames likely reduces noise that might otherwise interfere with model reasoning.

\subsection{Query-Adaptive Collaborative Inference}
\label{sec:confs-design}
\noindent \textbf{Shared Vision Representations.} Naively cascading model variants for collaboration incurs a penalty when the large VLM is eventually required. In such cases, the cumulative response delay can significantly exceed that of serving the query with the large model alone. To mitigate this, we explore optimization opportunities arising from the model architecture. VLMs are highly modular, which allows for the flexible integration or replacement of individual components. Notably, VLMs such as \emph{InstructBLIP} \bluecite{dai2023instructblip} have demonstrated that high performance can be achieved by reusing pretrained vision encoders and LMs without modification. In these cases, only the connector is fine-tuned for modality alignment. This architectural trait suggests that vision encoders can be decoupled from their specific LM counterparts. Motivated by this observation, we propose to share the entire video-to-token pipeline between the local and edge models. Ideally, this allows the edge-deployed large model to operate directly on the vision representations generated by the local pipeline. This sharing eliminates the overhead of re-executing the time-consuming video tokenization process.

To validate the feasibility of sharing vision encoders across VLM variants at inference time, we construct a hybrid model named \emph{VLM7B-Encoder2B}. This model is created by substituting the vision encoder of VideoChat-Flash-7B with the vision encoder of VideoChat-Flash-2B. This hybrid configuration incurs only a marginal accuracy drop compared to the native VideoChat-Flash-7B, with scores decreasing from 74.2 to 73.7 on MVBench and from 64.7 to 60.9 on Video-MME. Despite the small decreases, the hybrid model remains highly effective and noticeably outperforms the standalone 2B variant. Specifically, it improves accuracy from 69.9 to 73.7 on MVBench and from 57.4 to 60.9 on Video-MME. Although further aligning the shared vision encoder with both the 2B and 7B LMs via instruction tuning could theoretically bridge the small gaps, the hybrid model already exhibits sufficient accuracy without such retraining. Therefore, we adopt \emph{VLM7B-Encoder2B} as our large VLM in the following discussion. This choice enables our system to operate effectively while requiring minimal modifications to off-the-shelf models.

\noindent \textbf{Confidence-Based Routing.}
With the shared video tokenization pipeline minimizing the cost of offloading, the remaining challenge is determining \emph{when} to trigger edge augmentation. To maximize the effectiveness of the local-first strategy, the system must identify queries that the local model is unlikely to answer reliably. One might consider static \emph{pre-execution} routing based on text-only signals, such as predicting query complexity from the prompt\bluecite{dinghybrid}. Yet, such approaches cannot capture visual ambiguities like occlusion or motion blur that affect VLM performance. As a result, the Confidence-Based Router in QuickGrasp employs a \emph{post-execution} strategy that leverages the local VLM's inference confidence to determine whether edge augmentation is necessary.

To enable this post-execution routing, we first need quantitative measures of the model's uncertainty. Language modeling is inherently an autoregressive process, where the LM generates a sequence of tokens one step at a time. At each timestep $t$, the model predicts the probability distribution over the next token $y_t$ conditioned on the multimodal inputs (video and text) and the previously generated tokens $y_{<t}$. For closed-ended multiple-choice QA, the VLM is prompted to output the correct option letter (e.g., A-F) as its only generated token.\footnote{A sample system prompt is ``Answer with the option's letter from the given choices directly.''} Under this formatting, the uncertainty of the generated answer effectively collapses to the probability distribution over the set of valid option tokens at this initial generation step.

\begin{figure}[!t]
    \centering
    \includegraphics[width=\linewidth]{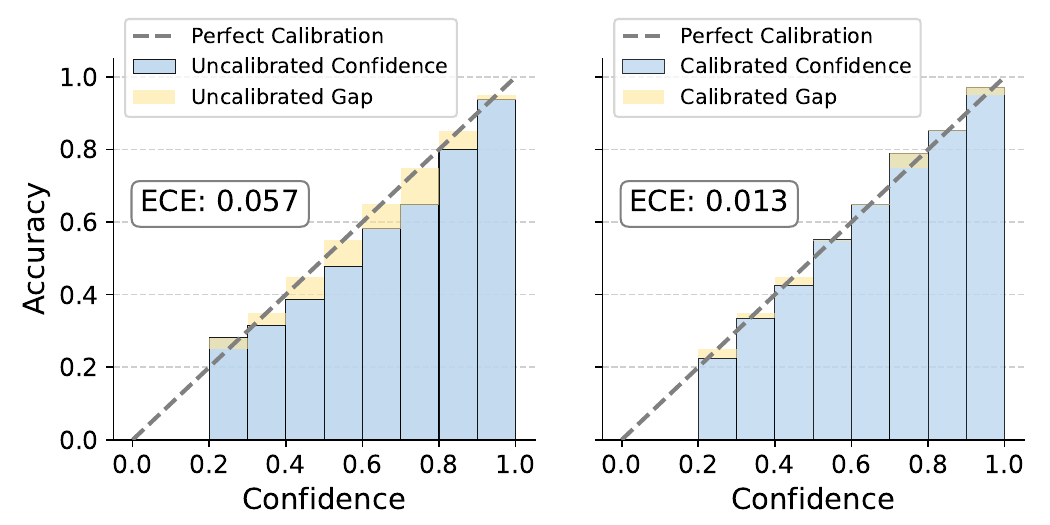}
    \caption{Confidence calibration of VideoChat-Flash-2B on MVBench (temperature scaling parameter $T$=$1.4$).}
    \label{fig:small_confidence}
\end{figure}

Formally, let $\mathbf{z}_i \in \mathbb{R}^{|\mathcal{V}|}$ denote the logit vector produced by the VLM at this specific generation step for query $i$, where $|\mathcal{V}|$ is the vocabulary size. Since the solution space is strictly constrained to the set of valid option letters $\mathcal{C}_i$ (e.g., $\mathcal{C}_i=\{$`A', $\cdots$, `F'$\}$), we are only concerned with the relative probability mass assigned to these tokens. We compute the predicted probability $p_{i,c}$ for each valid option $c \in \mathcal{C}_i$ by applying a constrained softmax function: 
\begin{equation} 
p_{i,c} = \frac{\exp(z_{i,c})}{\sum_{k \in \mathcal{C}_i} \exp(z_{i,k})} 
\end{equation} 
where $z_{i,c}$ is the logit component of $\mathbf{z}_i$, representing the raw logit value corresponding to token $c$ in the vocabulary. The model's output confidence is then defined as the maximum probability among all valid options $\max_{c \in \mathcal{C}_i} p_{i, c}$. This formulation effectively captures the model's conditional confidence when outputs are constrained to $\mathcal{C}_i$.

However, confidence scores from modern deep neural networks (DNNs) are often miscalibrated. The left subfigure in Fig.~\ref{fig:small_confidence} shows the reliability diagram\bluecite{guo2017calibration} of VideoChat-Flash-2B on the MVBench benchmark, where the predicted confidence of the model is generally higher than its empirical accuracy. For example, for predictions with confidence between 0.7 and 0.8, the empirical accuracy is 0.65. This overconfidence is consistent with previous findings that DNNs, including LMs, often produce probability estimates that overstate their true prediction accuracy\bluecite{guo2017calibration, wang2023tabi}.

To address this issue, the Confidence-Based Router adopts a standard confidence calibration technique, \emph{temperature scaling}\bluecite{guo2017calibration}. As a post-processing method, it requires no parameter updates or model retraining, preserving VLM modularity while incurring negligible computational overhead. Specifically, temperature scaling introduces a single positive scalar parameter $T$ to smooth the logit distribution before normalization. The calibrated confidence $\kappa_i$ for query $i$ is obtained by computing the constrained probability using the scaled logits $\mathbf{z}_i /T$:
\begin{equation} 
\kappa_i = \max_{c \in \mathcal{C}_i} \hat{p}_{i, c} =  \max_{c \in \mathcal{C}_i} \left( \frac{\exp(z_{i,c} / T)}{\sum_{k \in \mathcal{C}_i} \exp(z_{i,k} / T)} \right) 
\end{equation}
To quantify calibration quality, we report the \emph{expected calibration error (ECE)}\bluecite{guo2017calibration}, where lower values indicate better calibration. As shown in Fig.~\ref{fig:small_confidence}, applying temperature scaling reduces the ECE by approximately 77\% and narrows the gap between predicted confidence and empirical accuracy, demonstrating its effectiveness for VLMs.

Once calibrated, the local VLM confidence scores serve as the control signal for query-adaptive edge augmentation. Specifically, the Confidence-Based Router compares the calibrated confidence $\kappa_i$ against a predefined confidence threshold $\tau_{\mathrm{route}}$. When $\kappa_i$ falls below the threshold, the router treats the local model's response as unreliable and triggers edge augmentation. This design invokes augmentation only when needed, avoiding unnecessary overhead and maximizing the utility of the collaborative architecture.

\subsection{Delay-Aware and Accuracy-Preserving Configuration}
\label{sec:token-configuration}
\noindent \textbf{Token Density Formulation: Accuracy-Delay Trade-off.} Once the system triggers edge augmentation, the primary objective shifts to minimizing the network transmission and inference delay for the offloaded query. A key optimization lever lies in the connector of VLMs. Because LMs have a limited input context length, VLM connectors apply token reduction methods such as average pooling\bluecite{Maaz2024VideoChatGPT}, Q-Former\bluecite{li2023blip2}, or token merging\bluecite{bolya2023token} to compress the vision tokens passed to the LM. We refer to the average number of vision tokens retained per frame after this reduction process as \emph{token density}.

In general, lower token density reduces the volume of vision tokens transmitted and processed, which can decrease network transmission delay and edge inference delay. However, aggressive token reduction may degrade accuracy by discarding fine-grained visual details. To manage this trade-off, the Adaptive Token Configurator in QuickGrasp adaptively selects token density for each offloaded query. The objective is to use as few vision tokens as possible while maintaining high answer quality, thereby balancing result quality against response delay.

The long-term optimization objective of the Adaptive Token Configurator is:
\begin{equation}
    \argmax_{\{a_i \in \mathcal{A}\}} \sum_{i \in \mathcal{Q}_{edge}} \left( \mathds{1}(a_i) - \lambda \cdot (n_i \cdot a_i) \right)
\label{equ:opt-problem}
\end{equation}
where $\mathds{1}(a_i) \triangleq \mathds{1}\{\hat{y}_i^L(a_i)=y_i\}$.
Here, $\hat{y}_i^L(a_i)$ denotes the option predicted by the large VLM under token density $a_i$, and $y_i$ is the ground-truth option. $\mathds{1}\{\cdot\}$ is the indicator function that equals $1$ if the condition holds and $0$ otherwise. $n_i$ is the number of sampled frames for query $i$. The term $(n_i \cdot a_i)$ penalizes the offloaded vision token volume, and $\lambda > 0$ is a trade-off parameter balancing correctness against delay.

\noindent \textbf{Decision Framework: Contextual Multi-Armed Bandits.} Solving Eq.~(\ref{equ:opt-problem}) is challenging because changing token density affects accuracy in complex, non-linear ways that differ across queries. For instance, a simple query on a single‑object video (e.g., coarse‑grained action recognition) can often be answered correctly with a low vision token density, whereas a fine‑grained counting query on a complex scene normally requires high visual fidelity. As a result, static rules often fail to capture these content-dependent dynamics.

In our local-first serving architecture, the edge server processes offloaded queries sequentially to avoid resource contention. This yields an online per-query configuration selection problem with bandit feedback. For each query, the system observes the utility only for the selected configuration, while the utilities of unselected alternatives remain unobserved. Further, because each query is treated independently and does not affect the state of future queries, the problem does not involve cross-query state transitions. These properties naturally cast configuration selection as a CMAB problem.

In this framework, configuration selection is handled by an online learning agent. For each query $i$, the agent selects an action $a_i \in \mathcal{A}$ based on its context $\mathbf{c}_i$. It then observes a reward $r_i$ associated with the selected action and updates its \emph{action-selection policy} using this bandit feedback to maximize expected cumulative reward over time. To enable this context-aware decision making, we construct the context vector $\mathbf{c}_i$ by integrating the small VLM's calibrated uncertainty with the query's semantic features:
\begin{equation}
\mathbf{c}_i = \big[ n_i,\; \kappa_i^{S},\; \delta_i^{S},\; \mathcal{H}_i^{S},\; \mathbf{z}^{\mathrm{txt}}_{i},\; \mathbf{z}^{\mathrm{vis}}_{i},\; s_i^{\mathrm{max}},\; s_i^{\mathrm{mean}},\; s_i^{\mathrm{cplx}} \big]
\end{equation}
where $\kappa_i^S$, $\delta_i^{S}$, and $\mathcal{H}_i^{S}$ are derived from the calibrated confidence distribution of the small VLM over the valid options $\mathcal{C}_i$ for query $i$ (recall~\cref{sec:confs-design}). Specifically, $\kappa_i^{S}$ denotes the confidence of the top-ranked option (i.e., the maximum probability), $\delta_i^{S}$ is the margin between the top-1 and top-2 calibrated confidences, and $\mathcal{H}_i^{S}$ is the normalized entropy of the calibrated confidence distribution:
\begin{equation}
\mathcal{H}_i^{S} = \frac{-\sum_{c\in \mathcal{C}_i} \hat{p}_{i,c}\log \hat{p}_{i,c}}{\log |\mathcal{C}_i|}.
\end{equation}
These metrics collectively characterize local inference uncertainty and provide robust routing signals that are comparable across queries with different numbers of candidate options.

To efficiently capture the query's semantic context without incurring additional computational overhead, we directly reuse the intermediate embeddings generated during the small VLM inference. Let $D$ denote the hidden dimension of the small VLM's LM. We derive a condensed textual representation $\mathbf{h}^{\mathrm{txt}}_i \in \mathbb{R}^D$ by mean-pooling the question-token embeddings fed into the LM. Similarly, let $N$ represent the number of vision tokens passed to the LM. To obtain the visual representation $\mathbf{h}^{\mathrm{vis}}_{i} \in \mathbb{R}^D$, we extract the sequence of vision-token embeddings $\mathbf{H}^{\mathrm{vis}}_{i} \in \mathbb{R}^{N \times D}$ produced by the VLM's connector and aggregate them via global average pooling. 

However, directly concatenating these high-dimensional representations (e.g., $D=1536$) with scalar uncertainty features (e.g., $\kappa^S_i$) would likely cause them to dominate the decision process. To mitigate this while avoiding additional trainable parameters, we compress the pooled embeddings using principal component analysis (PCA). Specifically, we project $\mathbf{h}^{\mathrm{txt}}_i$ and $\mathbf{h}^{\mathrm{vis}}_i$ onto the top-$d$ principal components learned from the offline profiling:
\begin{equation}
    \mathbf{z}^{\mathrm{txt}}_i = \mathbf{P}_{t}^\top \mathbf{h}^{\mathrm{txt}}_i, \quad
    \mathbf{z}^{\mathrm{vis}}_i = \mathbf{P}_{v}^\top \mathbf{h}^{\mathrm{vis}}_i,
\end{equation}
where $\mathbf{P}_{t}, \mathbf{P}_{v} \in \mathbb{R}^{D \times d}$ are the PCA projection matrices and $d$ is a small latent dimension ($d=4$ by default). This design yields compact semantic features that are more balanced with the scalar uncertainty signals, and it reduces the risk of overfitting compared to training additional projection layers.

Finally, to explicitly measure the cross-modal relevance while mitigating the noise inherent in fine-grained token-level comparisons, we calculate cross-modal relevance metrics at the \emph{clip level}. We first compute the cosine similarity between the condensed textual representation $\mathbf{h}^{\mathrm{\mathrm{txt}}}_i$ and each vision-token embedding $\mathbf{H}^{\mathrm{\mathrm{vis}}}_{i,j}$. Instead of using these raw scores directly, we aggregate them temporally by summing the similarity scores of all vision tokens belonging to the same video clip. Assume the video consists of $N_i^{\mathrm{clip}}$ clips, and let $\mathcal{J}_t$ denote the set of token indices associated with the $t$-th clip. We then derive the clip-level relevance scores $R_{i,t}$ and the final aggregated features as follows:
\begin{equation}
\begin{aligned}
    s_{i,j} &= \frac{(\mathbf{h}^{\mathrm{txt}}_i)^\top \mathbf{H}^{\mathrm{vis}}_{i,j}}{\|\mathbf{h}^{\mathrm{txt}}_i\|_2 \|\mathbf{H}^{\mathrm{vis}}_{i,j}\|_2}, \\
    R_{i,t} &= \sum_{j \in \mathcal{J}_t} s_{i,j}, \quad \forall t \in \{1, \dots, N_i^{\mathrm{clip}}\} \\
    s_i^{\mathrm{max}} &= \max_{t} R_{i,t}, \quad \quad s_i^{\mathrm{mean}} = \frac{1}{N_i^{\mathrm{clip}}} \sum_{t=1}^{N^{\mathrm{clip}}_i} R_{i,t}
\end{aligned}
\end{equation}
where $R_{i,t}$ represents the cumulative semantic alignment of the $t$-th video clip with the textual question. By aggregating at the clip level, $s_i^{\mathrm{max}}$ and $s_i^{\mathrm{mean}}$ effectively capture the relevance of the most pertinent video clips and the overall video context, respectively, providing a robust signal for decision making.

To further gauge the visual information density required to answer the query, we introduce a \emph{query-grounded spectral complexity} metric, $s_i^{\mathrm{cplx}}$. We first identify the most relevant clips based on their relevance scores $R_{i,t}$. Specifically, among the $N_i^{\mathrm{clip}}$ clips for query $i$, we select the top $\min(N_i^{\mathrm{clip}}, K)$ clips with the largest $R_{i,t}$ values and denote their indices as $\mathcal{K}_i$, where $K=3$ by default. For each selected clip $t \in \mathcal{K}_i$, we form its vision-token matrix $\mathbf{X}_t \in \mathbb{R}^{N_{\mathrm{tok}}\times D}$, where each row corresponds to one token embedding of dimension $D$ and $N_{\mathrm{tok}}$ is the number of tokens in the clip. We then compute the singular value decomposition (SVD) within the clip:
\begin{equation}
    \mathbf{X}_t = \mathbf{U}_t \bm{\Sigma}_t \mathbf{V}_t^\top,
\end{equation}
and obtain the singular values $\{\sigma_{t,j}\}_{j=1}^{J}$, where $J=\min(N_{\mathrm{tok}},D)$. We normalize them to a probability distribution $p_{t,j}$ and compute the spectral entropy:
\begin{equation}
    H_t = -\sum_{j=1}^{J} p_{t,j}\log p_{t,j}.
\end{equation}
The final complexity score is the average normalized entropy over the selected clips:
\begin{equation}
    s_i^{\mathrm{cplx}} = \frac{1}{|\mathcal{K}_i|} \sum_{t \in \mathcal{K}_i} \frac{H_t}{\log J},
\end{equation}
where the normalization term $\log J$ ensures $s_i^{\mathrm{cplx}} \in [0,1]$. A high $s_i^{\mathrm{cplx}}$ indicates a flatter singular-value spectrum, suggesting that the semantically relevant clips are visually complex and may require a higher token density for accurate reasoning.


To align strictly with the long-term optimization objective in Eq.~(\ref{equ:opt-problem}), we define the reward $r_i$ for offloading query $i$ to the edge with token density $a_i$ as follows:
\begin{equation} 
r_i = \mathds{1}(a_i) - \lambda \cdot (n_i \cdot a_i) 
\label{equ:reward}
\end{equation}
Here, the agent receives a unit reward for a correct prediction and is penalized by the weighted cost $\lambda \cdot (n_i \cdot a_i)$. This structure ensures that maximizing the expected reward ultimately leads to a policy that prioritizes efficiency, favoring higher token densities only when strictly necessary to convert a wrong answer into a correct one. For any given query $i$, once an action $a_i$ is proposed, the associated penalty term $\lambda \cdot (n_i \cdot a_i)$ is deterministic and known. Consequently, the agent's primary challenge is to accurately estimate the correctness probability for each (context, action) pair.

\noindent \textbf{Proposed Solution: Neural Linear Bandit.} 
Due to the complex, non-linear dependency of VLM answer correctness on token density, we propose a \emph{neural linear bandit} approach. Specifically, we employ an MLP architecture where the hidden layers act as a deep feature extractor $\bm{u}_i = \phi_{\theta}(\mathbf{c}_i)$. This backbone transforms the query context $\mathbf{c}_i$ into a latent representation $\bm{u}_i \in \mathbb{R}^{d_u}$. During the offline profiling phase, we attach a multi-head output layer containing $|\mathcal{A}|$ nodes to this backbone, allowing it to predict the correctness probability for each candidate action $a \in \mathcal{A}$ simultaneously. The rationale for this design is twofold. First, the multi-head training objective forces the latent representation $\bm{u}_i$ to capture high-dimensional cross-modal interactions and semantic features that are most discriminative for predicting VLM performance across different token densities. Second, this architecture allows the system to share learned knowledge about the video content across all possible actions within a single forward pass, significantly improving sample efficiency.

To address the cold-start problem and ensure high initial performance, we train the backbone parameters $\theta$ offline using a curated \emph{profiling dataset}, $\mathcal{D}_\mathrm{prof}=\{(\mathbf{c}_i, a_i, \mathds{1}(a_i))\}_{i=1}^m$, constructed from ground-truth labeled queries. After offline training, we freeze $\theta$, yielding a stable and robust latent space for online operations. We then discard the training heads and instead attach a set of contextual linear bandits, maintaining an independent Bayesian linear regressor for each action on top of the latent representation. One might argue for directly updating the MLP online and using MC-Dropout to approximate Thompson sampling \bluecite{gal2016dropout}. However, MC-Dropout often yields miscalibrated uncertainty and requires multiple forward passes to stabilize, which both increases decision delay and still falls short of providing reliable uncertainty for exploration. By decoupling feature extraction from decision making, the linear bandit maintains a principled Bayesian posterior with well-defined predictive uncertainty, enabling reliable reward estimation and efficient exploration via Thompson sampling.

\noindent \textbf{Online Adaptation: Exploration and Policy Update.} 
In a live serving environment, ground-truth answers are typically unavailable, making the true correctness in Eq.~(\ref{equ:reward}) unobservable. We therefore rely on the calibrated confidence 
$\kappa_i^L$ of the large VLM as a reliability proxy and define an online proxy reward $\tilde{r_i}$ as
\begin{equation}
    \tilde{r_i} = \mathds{1}^{\mathrm{proxy}}(a_i) - \lambda \cdot (n_i \cdot a_i)
    \label{equ:proxy-reward}
\end{equation}
where $\mathds{1}^{\mathrm{proxy}}(a_i) \triangleq \mathds{1}\{\kappa_i^L > \tau_{\mathrm{proxy}}\}$. $\tau_{\mathrm{proxy}}$ is a confidence threshold. This reward structure implements a self-training mechanism where the term $\mathds{1}\{\cdot\}$ acts as a reliability filter, treating high-confidence predictions as ground-truth surrogate labels. By assigning a unit correctness reward only when the VLM is highly certain, we encourage the policy to select the most efficient token density that satisfies the rigorous reliability constraint.

To update the policy online under the proxy reward in Eq.~(\ref{equ:proxy-reward}), we perform a per-interaction Bayesian update of the action-wise linear bandit. Although $\mathds{1}^{\mathrm{proxy}}$ is binary, we use Bayesian linear regression as a lightweight surrogate to obtain uncertainty estimates for Thompson sampling. The resulting predictor produces a real-valued score, which we interpret as a success probability by clipping it to $[0,1]$. For each action $a$, the system maintains the Bayesian linear regression sufficient statistics $A_a \in \mathbb{R}^{d_u \times d_u}$ and $b_a \in \mathbb{R}^{d_u}$, where $A_a$ accumulates feature outer products and $b_a$ accumulates reward-weighted features. We initialize these sufficient statistics using the profiling dataset $\mathcal{D}_{\mathrm{prof}}$ to warm-start the posterior and mitigate cold-start exploration. Specifically, we set $A_a \leftarrow \lambda_0 I$ and $b_a \leftarrow \mathbf{0}$, and then accumulate all offline samples in $\mathcal{D}_{\mathrm{prof}}$ whose action equals $a$:
\begin{equation}
\begin{aligned}
    A_a &\leftarrow \lambda_0 I + \sum_{(\mathbf{c}_j, a_j, \mathds{1}(a_j)) \in \mathcal{D}_{\mathrm{prof}} \mid a_j=a} \bm{u}_j \bm{u}_j^\top, \\
    b_a &\leftarrow \sum_{(\mathbf{c}_j, a_j, \mathds{1}(a_j)) \in \mathcal{D}_{\mathrm{prof}} \mid a_j=a} \bm{u}_j \, \mathds{1}(a_j),
\end{aligned}
\label{equ:bandit-warmup}
\end{equation}
where $\bm{u}_j = \phi_{\theta}(\mathbf{c}_j)$ denotes the frozen latent feature produced by the pretrained backbone. During online serving, after each interaction we obtain a new experience $(\bm{u}_i, a, \mathds{1}^{\mathrm{proxy}}(a))$ and update the sufficient statistics of the selected action as:
\begin{equation}
\begin{aligned}
    A_a &\leftarrow A_a + \bm{u}_i \bm{u}_i^\top, \\
    b_a &\leftarrow b_a + \bm{u}_i \mathds{1}^{\mathrm{proxy}}(a)
\end{aligned}
\label{equ:bandit-update}
\end{equation}
The posterior mean is $\beta_a = A_a^{-1} b_a$. During inference, the agent performs Thompson sampling by drawing a realization $\hat{\beta}_a \sim \mathcal{N}(\beta_a, \alpha^2 A_a^{-1})$ for each action and forming the clipped success-probability estimate
\begin{equation}
    \hat{p}_i(a) = \mathrm{clip}\!\left(\bm{u}_i^\top \hat{\beta}_a,\,0,\,1\right),
\label{equ:clipped-prob}
\end{equation}
which is then used to select the action
\begin{equation}
    a_i = \arg \max_{a \in \mathcal{A}} \left( \hat{p}_i(a) - \lambda \cdot (n_i \cdot a) \right).
\end{equation}
This per-interaction update keeps the policy responsive while avoiding any retraining overhead in the inference pipeline.
\section{Service Prototype Implementation}
\label{sec:prototype}

We deploy the QuickGrasp prototype on a representative device-edge collaborative testbed to evaluate its performance in real-world service scenarios.

\noindent \textbf{Service Infrastructure.} The edge service node acts as the remote inference backend and is provisioned with an NVIDIA RTX Pro 6000 Blackwell GPU (96 GB VRAM), an Intel(R) Core(TM) Ultra 7 265K CPU, and 64 GiB of RAM. The local service node represents a modern AI-capable endpoint and is configured with an NVIDIA GeForce RTX 4090 GPU (24 GB VRAM), an Intel(R) Core(TM) i9-14900K CPU, and 64 GiB of RAM. Both service nodes operate on the Ubuntu 24.04 LTS operating system.

\noindent \textbf{Network Environment Emulation.} To ensure fair comparisons and reproducibility, we rigorously control the communication link between the local and edge nodes using the Linux traffic control (\texttt{tc}) utility\bluecite{tc}. For the default experimental configuration, we shape the upload throughput from the local node to the edge node to 59 Mbps, the download throughput to 119 Mbps, and the round-trip time (RTT) to 9 ms. These parameters are selected to mirror the median global network performance reported by Speedtest\bluecite{speedtest}.

\noindent \textbf{Software and Distributed Service Orchestration.} The complete service orchestration logic is implemented in \texttt{Python} with all DNN components executed via \texttt{PyTorch}. For video data processing, we utilize the \texttt{Decord} library as it ensures high-throughput frame sampling which is a critical enabler for our accelerated tokenization pipeline. Device-edge communication is governed by \texttt{ZeroMQ}\bluecite{zmq} and was selected for its lightweight asynchronous messaging capabilities and minimal protocol overhead. To maximize bandwidth efficiency during offloading, we employ a rigorous data optimization strategy where vision tokens are losslessly compressed using \texttt{Zstandard}\bluecite{zstandard} before transmission. This library is chosen for its favorable trade-off between compression ratio and throughput. Additionally, we transmit the raw textual question rather than pre-tokenized sequences, as the integer-based token representation significantly increases payload size.

\section{Evaluation}
\label{sec:evaluation}
\subsection{Evaluation Setup}
\label{sec:eva-setup}

\noindent \textbf{Benchmarks:} To ensure a comprehensive evaluation, we use three recently proposed benchmarks for multimodal video understanding\bluecite{li2024mvbench, fu2025videomme, zhou2025mlvu}. These benchmarks cover videos across diverse genres, resolutions, frame rates, bitrates, and durations. The associated tasks require a range of spatiotemporal reasoning capabilities, such as temporal reasoning (e.g., action prediction), object reasoning (e.g., object shuffling), scene understanding (e.g., scene transitions), attribute perception (e.g., state changes), and counting (e.g., object motion counting).

\begin{itemize}[label=$\diamond$, leftmargin=*]
    \item \emph{MVBench}\bluecite{li2024mvbench} comprises 3,800 queries (question-answer pairs) across 19 temporally grounded tasks.\footnote{The reported numbers are obtained after excluding videos provided as image sequences. To better reflect practical serving scenarios, we consider only videos provided in encoded video formats (e.g., MP4, MKV, AVI).} The benchmark consists primarily of short clips (duration $<$ 35 seconds), making it ideal for evaluating the system's responsiveness on atomic events. Each question is accompanied by up to 5 candidate options.
    
   \item \emph{Video-MME}\bluecite{fu2025videomme} includes 2,700 queries spanning 6 distinct visual domains. It features videos with durations spanning from 11 seconds to 1 hour, testing the system's adaptability to varying temporal contexts. Each question provides 4 candidate options.
   
    \item \emph{MLVU-Test}\bluecite{zhou2025mlvu} focuses on long video understanding. We utilize its test set, which poses a distinct challenge with extended video durations ranging from 187 seconds to over 2 hours (8,206 seconds). This set covers 9 task types and features a more difficult selection format with 6 candidate options per question, rigorously testing the system's ability to reason across extended temporal windows.
\end{itemize}

\noindent \textbf{VLM Models.} We select VideoChat-Flash as the representative model family for our evaluation. Specifically, we deploy the pre-trained 2B variant\bluecite{videochat-2b} on the local node and the pre-trained 7B variant\bluecite{videochat-7b} on the edge node. Unless otherwise stated, we utilize the default configuration parameters provided by the model authors, such as sampling $4$ frames per video clip. To minimize processing delay, we optimize the decoding and preprocessing pipelines of the original implementation by enabling multi-threaded decoding and batch preprocessing. Crucially, as our contribution lies in system-level optimization and distributed service orchestration rather than VLM architecture design, we do not fine-tune or modify the weights of the pre-trained VLMs. Although minimal interface adaptations may be required for other architectures, our system is designed to be model-agnostic and applicable to a wide range of modern VLM families.

\begin{figure*}[!t]
    \centering
    \includegraphics[width=\linewidth]{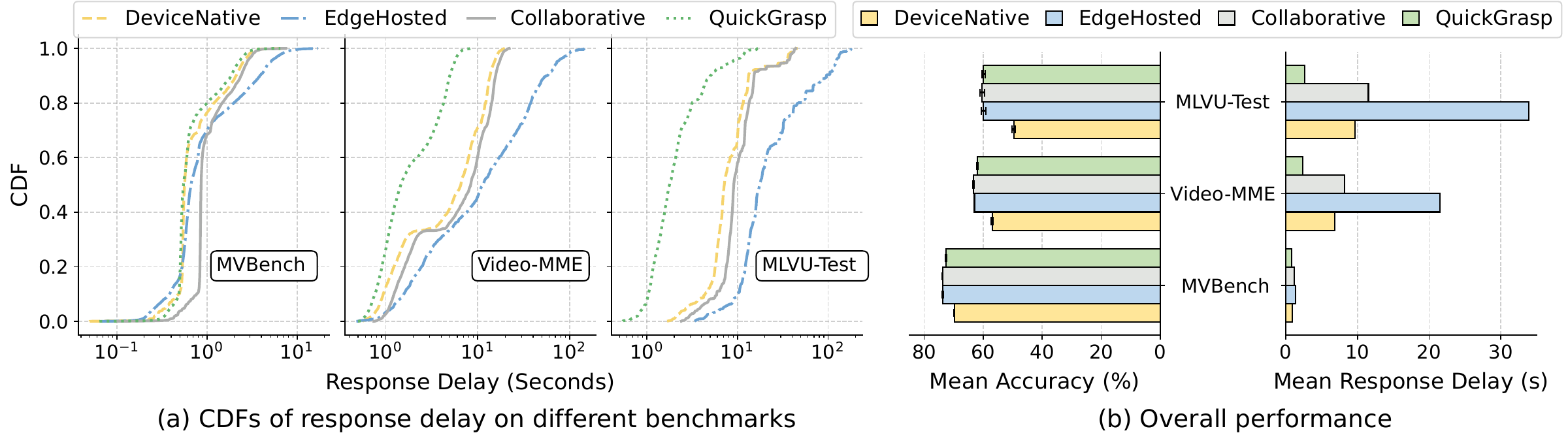}
    \caption{Performance comparison of different solutions on distinct benchmarks. (Results are aggregated over five independent runs per solution to account for randomness.)}
    \label{fig:expt-res-basic}
\end{figure*}

\noindent \textbf{Methodology.} We randomly partition each benchmark dataset into a profiling set and a test set using a 3:7 ratio. The profiling set is utilized to derive the temperature scaling parameters for the confidence calibration of both small and large VLMs and to initialize the neural linear bandit. To ensure strict service fidelity, we determine the local routing threshold $\tau_{\mathrm{route}}$ and the large VLM proxy reward threshold $\tau_{\mathrm{proxy}}$ from the profiling set, setting both to 0.6. The hyperparameters of the neural feature extractor are optimized using Optuna\bluecite{optuna}, resulting in an MLP backbone comprising two hidden layers (128 and 64 units) and a final latent dimension of $d_u=64$. For the contextual linear bandits, we set the prior precision parameter $\lambda_0 = 0.1$ and the Thompson sampling exploration coefficient $\alpha = 0.05$. To ensure that the accuracy and cost penalty remain on a comparable scale, we normalize the cost term by configuring the trade-off weight $\lambda$ such that the maximum possible token volume scales to 1.0. We consider five discrete candidates for an action $a$, defined as $\mathcal{A}=\{2, 4, 8, 16, 32\}$.

\noindent \textbf{Baselines.} To evaluate the effectiveness of our solution, we compare QuickGrasp against three representative solutions:
\begin{itemize}[label=$\diamond$, leftmargin=*]
    \item \emph{DeviceNative.} This solution aligns with the active research direction of lightweight VLM design \bluecite{Shaker2025MobileVideoGPT, marafioti2025smolvlm}, enabling a standalone service provisioning strategy. In this paradigm, the VLM is hosted exclusively on the end device to ensure strict service availability without network dependence. We instantiate this strategy using a locally deployed VideoChat-Flash-2B model to serve queries.

    \item \emph{EdgeHosted.} This solution implements the canonical computation offloading paradigm\bluecite{gpt-4o, google-gemini}. The local node operates as a thin client, transmitting the original user query (comprising the textual question and raw video file) to the edge node. The edge node hosts the VideoChat-Flash-7B model to service the incoming queries.

    \item \emph{Collaborative.} Following existing collaborative inference paradigms \bluecite{jin2025collm, jiang2025energy}, this baseline partitions the VideoChat-Flash-7B model between local and edge nodes. To minimize transmission overhead, the local node handles video tokenization and sends only the textual question and compressed vision tokens to the edge node, which then executes the memory-intensive LLM inference to generate responses.
    
\end{itemize}

\noindent \textbf{Evaluation Metrics:} We primarily consider \emph{accuracy} and \emph{response delay}. Accuracy is defined as the proportion of correctly answered queries for each benchmark. Response delay serves as our primary QoS metric, defined from the user's perspective as the total elapsed wall-clock time from the moment a query is submitted to the moment the final answer is received.

\begin{figure}[!t]
    \centering
    \includegraphics[width=\linewidth]{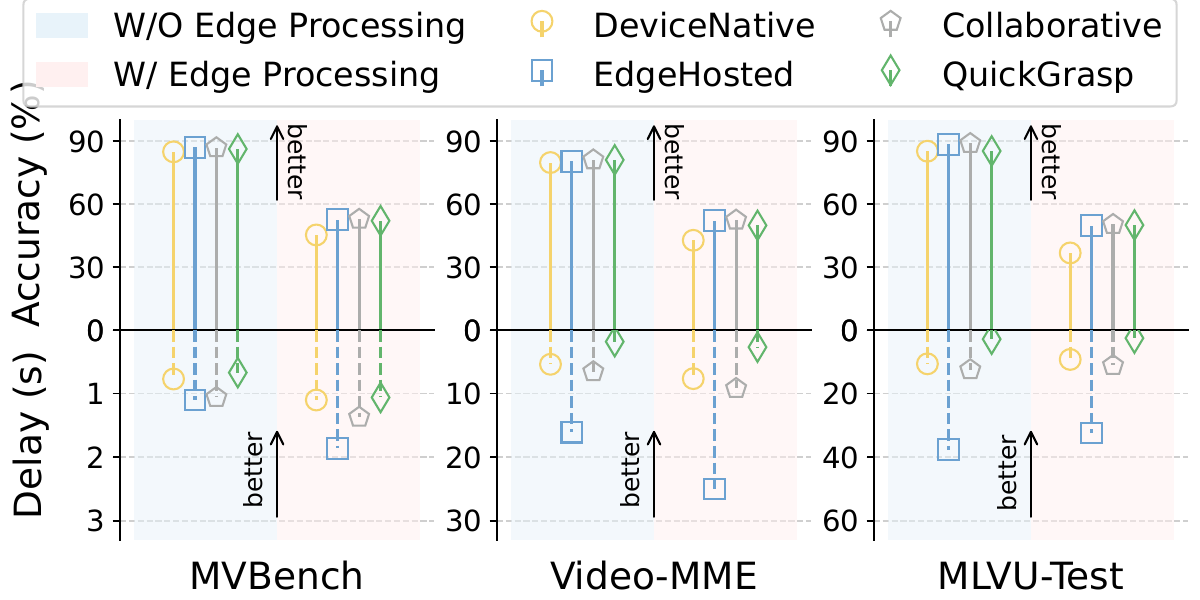}
    \caption{Performance comparison of queries answered with (w/) and without (w/o) edge processing.}
    \label{fig:expt-local-edge}
\end{figure}

\subsection{Overall Performance Improvements}

Approximately 20\% of the videos in MVBench have a duration under 5 seconds and a compressed file size below 821 KiB. For these short, low-quality videos, EdgeHosted achieves the lowest response delay and even outperforms DeviceNative and QuickGrasp. This occurs because the large VLM inference speed on the server node for such short videos is slightly faster than the small VLM inference speed on the local device. The minimal network transmission cost for these small files is effectively offset by this server-side processing advantage. In contrast, even with compression, the intermediate vision tokens transmitted by Collaborative can be larger than the original compressed video. This data amplification makes token transmission slower than uploading the compressed video, explaining why Collaborative incurs higher response delay than EdgeHosted for 73\% of MVBench queries. As video duration increases, the size of compressed video files scales more aggressively than the vision tokens, making tokenized representations increasingly transmission-efficient. This fundamental scaling gap explains why EdgeHosted consistently incurs the highest response delay on the longer videos of Video-MME and MLVU-Test.

Overall, QuickGrasp achieves mean accuracies significantly higher than DeviceNative and comparable to the edge-centric baselines, namely EdgeHosted and Collaborative, while markedly reducing mean response delays. As presented in Fig. \ref{fig:expt-res-basic}(b), the mean response delay of QuickGrasp on MVBench is approximately 819 ms. This is lower than the fastest baseline, DeviceNative, which records approximately 900 ms. This speed advantage is primarily driven by accelerated video tokenization, which remains effective even for the short videos in MVBench. On Video-MME across varying video lengths, QuickGrasp delivers 2.9$\times$, 3.5$\times$, and 9.1$\times$ speedups over DeviceNative, Collaborative, and EdgeHosted, respectively. On MLVU-Test, which contains much longer videos, the speedup is even more pronounced. It reaches 12.8$\times$ relative to EdgeHosted and 4.4$\times$ relative to Collaborative with only a slight decrease in mean accuracy from 60.3 to 59.9. The CDFs in Fig.~\ref{fig:expt-res-basic}(a) further illustrate this consistency, showing that QuickGrasp maintains a tight distribution of low response delays regardless of video length.

\subsection{In-depth Analysis}

\begin{figure*}
     \begin{minipage}[t]{0.69\linewidth}
        \centering
        \includegraphics[width=\linewidth]{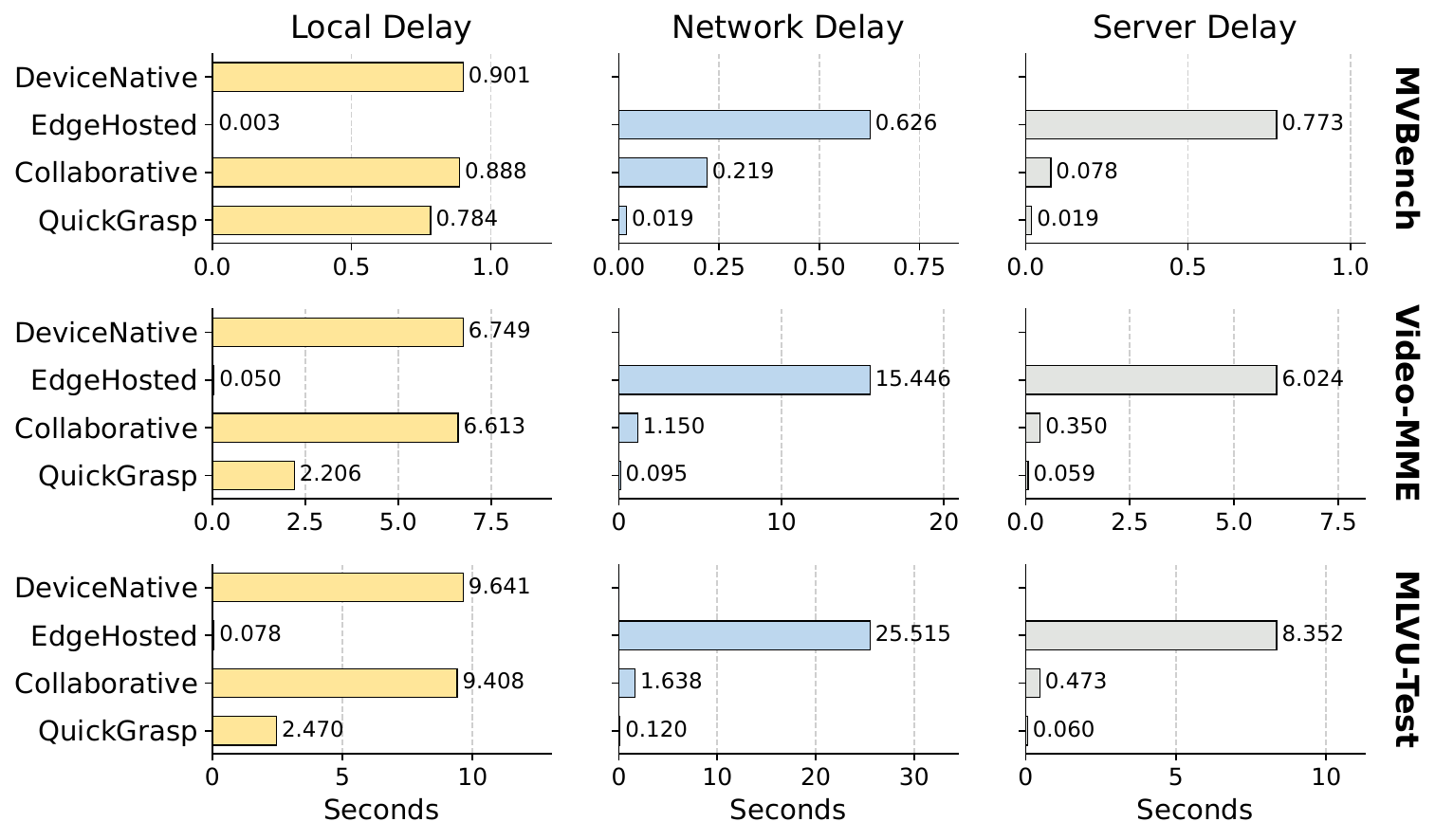}
        \caption{Response delay breakdown. The reported values are averaged over all queries in the corresponding benchmark.}
        \label{fig:expt-delay-breakdown}
    \end{minipage}
    \hfill
    \begin{minipage}[t]{0.27\linewidth}
        \centering
        \includegraphics[width=\linewidth]{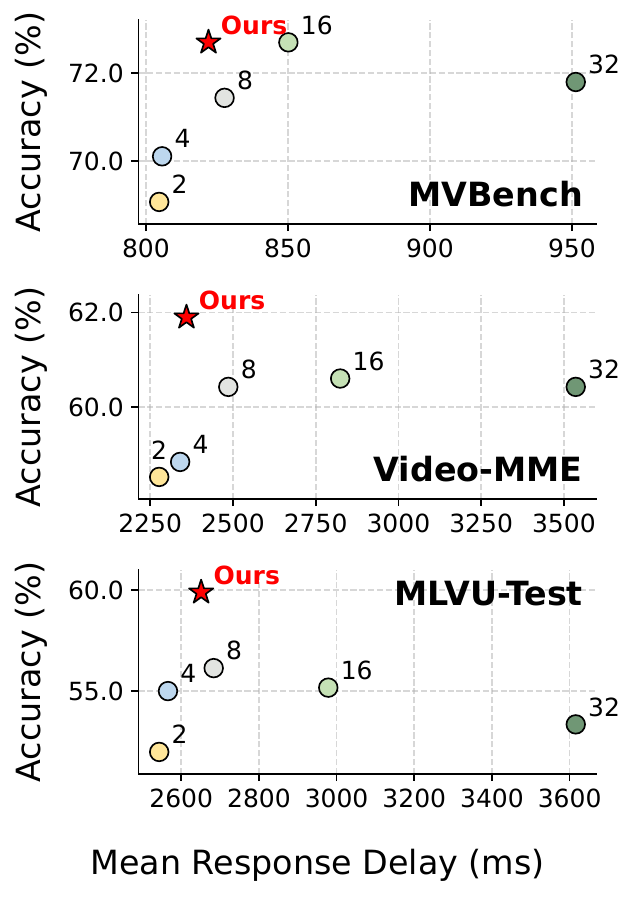}
        \caption{Ablation study comparing fixed token densities against our approach.}
        \label{fig:expt-ab1-fixed-td}
    \end{minipage}
    \vspace{-0.3cm}
\end{figure*}

\textbf{Performance Breakdown by Offloading Decision.} Based on the offloading decisions made by QuickGrasp, we categorize queries into two subsets: those processed entirely on the local device and those offloaded to the edge for augmentation. We compare the performance of different solutions on these two subsets in Fig.~\ref{fig:expt-local-edge}. For queries that QuickGrasp determines can be answered locally, the achieved accuracy remains high and closely matches that of the edge-centric baselines. This suggests that edge processing offers little to no additional benefit for these queries, demonstrating QuickGrasp's ability to identify and retain easier queries for local execution. By handling these queries locally, QuickGrasp achieves response delay speedups ranging from 1.2$\times$ to 13.1$\times$.

For queries that QuickGrasp identifies as challenging and offloads to the edge, it achieves substantial accuracy improvements over DeviceNative. Although these queries involve network transmission, the associated response delays remain comparable to those of DeviceNative on short videos and significantly lower on longer videos due to accelerated video tokenization. Relative to other network-based baselines, QuickGrasp maintains comparable accuracy while markedly reducing response delay, yielding 1.3$\times$ to 12.6$\times$ speedups.


\textbf{Communication and edge processing costs.} We utilize network communication delay to quantify communication cost, as it captures the holistic effects of both network conditions and the volume of data transmitted. Similarly, we use server processing delay to represent edge processing cost. Fig.~\ref{fig:expt-delay-breakdown} demonstrates the breakdown of response delays for different solutions across the benchmarks. As expected, DeviceNative incurs no communication or edge processing costs since all computation is performed locally. In contrast, EdgeHosted bears the highest costs due to its fully offloaded strategy. Collaborative significantly reduces both communication and server processing costs by offloading only part of the workload, achieving 2.9$\times$ to 15.6$\times$ network speedup and 10.0$\times$ to 17.7$\times$ server processing speedup compared to EdgeHosted. With optimized frame sampling and video tokenization, QuickGrasp achieves lower local processing delays than DeviceNative and Collaborative. Moreover, by selectively offloading vision tokens in a delay-aware and accuracy-preserving manner, it significantly reduces network communication and edge processing delays compared to edge-centric approaches. These results highlight the cost efficiency of our solution.

\subsection{Ablation Study and System Overhead}

\begin{figure}
     \begin{minipage}[t]{0.475\linewidth}
        \centering
        \includegraphics[width=\linewidth]{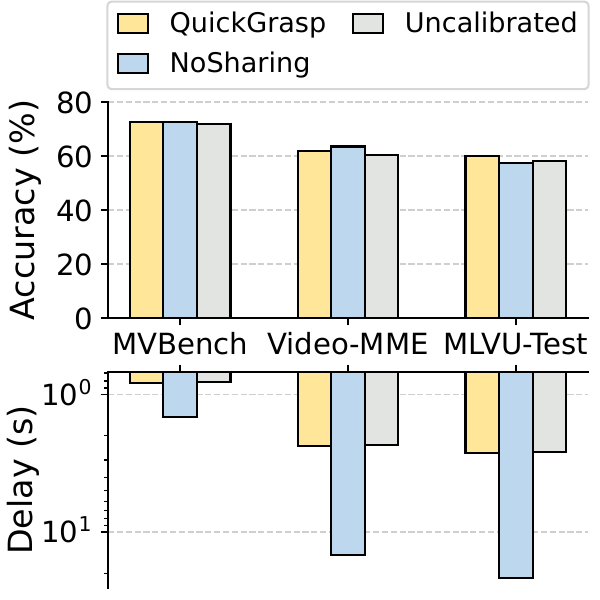}
        \caption{Ablation study.}
        \label{fig:ablation-study}
    \end{minipage}
    \hfill
    \begin{minipage}[t]{0.475\linewidth}
        \centering
        \includegraphics[width=\linewidth]{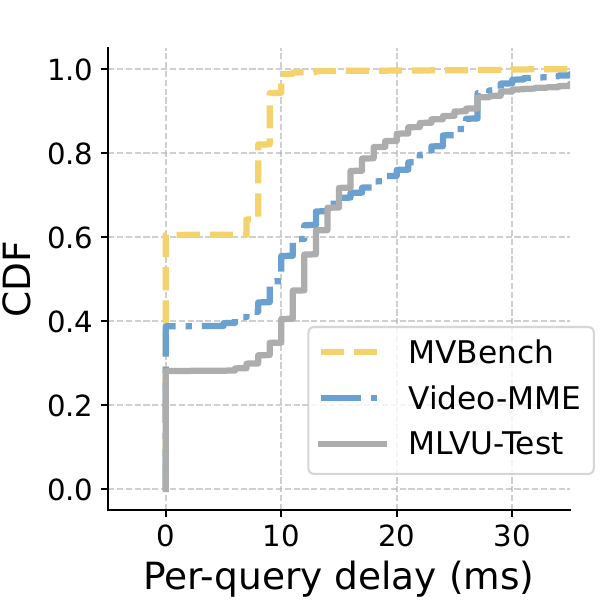}
        \caption{System overhead.}
        \label{fig:system-overhead}
    \end{minipage}
    \vspace{-0.3cm}
\end{figure}

\textbf{Ablation Study.} We first evaluate the effectiveness of our adaptive token density configuration by comparing it against ablated versions that use a fixed token density from the set $\mathcal{A}$. The results are presented in Fig.~\ref{fig:expt-ab1-fixed-td}. Across all three benchmarks, our adaptive solution significantly outperforms the fixed configurations by achieving a superior accuracy and delay trade-off. As expected, increasing the token density from 2 to 32 strictly increases the mean response delay, which aligns with the higher computation and transmission costs. Accuracy, however, does not scale monotonically. Instead, it follows an inverted-U trend, peaking at intermediate densities (e.g., 16 for MVBench and Video-MME, and 8 for MLVU-Test) before degrading at higher densities. This trend suggests that while too few vision tokens discard informative visual content, excessive tokens introduce redundant or noisy signals that distract the large VLM and hinder its reasoning capabilities. By contrast, our adaptive approach intelligently customizes the token density on a per-query basis. This allows QuickGrasp to surpass the peak accuracy of the best fixed configuration on every benchmark, while simultaneously maintaining a mean response delay comparable to the relatively sparse settings (e.g., density 8). These results clearly demonstrate the necessity and superiority of dynamic density configuration in balancing semantic richness with service responsiveness.

Fig.~\ref{fig:ablation-study} further compares the full version of QuickGrasp with two ablated variants. The first variant, \emph{NoSharing}, disables vision representation sharing and treats the local and edge VLMs as independent standalone models. Consequently, the adaptive token density configuration is also disabled. Once edge augmentation is required, the entire query is transmitted to the edge for answer generation. The second variant, \emph{Uncalibrated}, removes the confidence calibration module for both the local and edge VLMs, relying instead on uncalibrated confidence scores to drive offloading decisions.

As shown in Fig.~\ref{fig:ablation-study}, although the accelerated video tokenizer is enabled for both local and edge inference, the NoSharing variant exhibits significantly higher response delays, with comparable accuracy to QuickGrasp on MVBench and a slightly lower accuracy on MLVU-Test, while achieving higher accuracy on Video-MME. This added delay is mainly due to transmitting the original query (i.e., video plus question) and redundant video tokenization on the server side. On MVBench, the contributions of these two factors are comparable, with original query transmission and redundant tokenization accounting for approximately 49\% and 51\% of the total delay increase, respectively. On Video-MME and MLVU-Test, the delay increase is dominated by original-query transmission because of the longer videos, accounting for 89\% and 92\% of the total added delay, respectively.

By contrast, the Uncalibrated variant attains response delays similar to QuickGrasp but incurs noticeable accuracy drops. This degradation mainly comes from errors in locally served queries. Without calibration, the local VLM becomes overconfident, so more incorrect predictions exceed the acceptance threshold and are returned directly instead of being corrected through edge augmentation. We also observe a slight accuracy drop on offloaded queries, which we attribute to suboptimal token density selection. A likely reason is that an overconfident large VLM provides unreliable feedback during online adaptation, causing some incorrect actions to receive a unit correctness reward and biasing the policy updates.

\textbf{System Overhead.} The system overhead of QuickGrasp primarily arises from the configuration selection and vision token compression for offloading. Fig.~\ref{fig:system-overhead} plots the CDFs of the per-query delays incurred by these operations across all benchmarks. Owing to the delay-aware design, this overhead remains small, typically within tens of milliseconds. Consequently, it is negligible relative to the E2E response delay, accounting on average for only 0.46\%, 0.50\%, and 0.67\% on MVBench, Video-MME, and MLVU-Test, respectively.

\section{Related Work}
\label{sec:related-work}

\textbf{Resource-Efficient Multimodal Model Serving.} As multimodal LLMs are increasingly deployed in latency-sensitive applications, recent work has focused on improving serving efficiency from both model-architectural and system-scheduling perspectives. To mitigate the substantial computational cost induced by high-dimensional visual inputs, many efforts have been made to reduce the vision token volume processed by the LM. FastV\bluecite{chen2024image} finds that the attention paid to vision tokens is substantially lower in the deep layers of the LLM than in the first few layers. It then prunes a fraction of vision tokens after an early LLM layer to reduce computation and LLM generation delay with minimal accuracy loss. DyCoke\bluecite{tao2025dycoke} reduces generation overhead by merging temporally redundant vision tokens at the prefilling stage and removing spatially redundant vision tokens dynamically at the decoding stage. NVILA\bluecite{liu2025nvila} proposes a training-aware scale-then-compress paradigm that applies static spatial pooling and temporal averaging to reduce the number of tokens fed into the LM. While these intra-model optimizations effectively accelerate the LM generation phase, they overlook the severe bottlenecks inherent in video tokenization. Moreover, these single-node designs inherently ignore the communication delays and offloading trade-offs of device-edge collaboration.

Beyond the model itself, system-level efforts optimize how multimodal inference stages map to heterogeneous resources. EPD Disaggregation\bluecite{singh2025efficiently} and ModServe\bluecite{qiu2025modserve} advocate disaggregated serving by decoupling the multimodal encoder from language inference to reduce contention and enable stage-specific scheduling and scaling. Specifically, EPD Disaggregation separates encoding, LLM prefilling, and decoding onto dedicated GPU pools, whereas ModServe groups the pipeline into image instances for preprocessing and encoding and text instances for LLM prefill and decoding. The disaggregation mitigates cross-stage interference and improves cluster throughput and efficiency. Unlike these frameworks that optimize multi-node clusters, Nova\bluecite{xu2025nova} targets single-GPU efficiency for agentic vision-language model serving, using adaptive cross-stage pipeline parallelization with fine-grained streaming multiprocessor partitioning to maintain real-time responsiveness. However, whether scaling out across a cluster or optimizing within a single GPU, these solutions typically assume centralized, high-bandwidth hardware. They do not address the joint resource constraints of device-edge collaborative environments.

\textbf{QoS-Aware Collaborative Inference Services.} Collaborative device-edge or edge-cloud inference has emerged as a common paradigm to extend the capabilities of resource-constrained local devices. Prior device-edge collaborative inference often partitions a DNN at a cut layer, executes the preceding layers on the device, and offloads the remaining layers to an edge server, with the partition point and resource allocation optimized to meet energy, latency, and accuracy goals\bluecite{jiang2025energy}. Incentive-aware mechanisms have also been explored for collaborative inference across multiple distributed edge devices, jointly optimizing partitioning and scheduling decisions\bluecite{kim2024incentive}. Beyond single-shot partitioning, sequential applications create cross-task dependencies that couple offloading decisions over time. Teng et al.\bluecite{teng2025integrated} address this by jointly optimizing the offloading breakpoints and multi-resource allocation to reduce overall system cost. With the rise of generative AI, more recent work addresses the unique challenges of LLMs. In this context, cloud-edge scheduling frameworks leverage combinatorial bandits to predict the processing time of unpredictable LLM requests and dynamically route them to mitigate queuing bottlenecks\bluecite{li2025cloud}.

QuickGrasp advances the collaborative paradigm by introducing a content-aware orchestration framework tailored specifically for VLMs. Unlike generic DNN partitioning that splits sequential layers, QuickGrasp exploits the modularity of VLMs to share vision representations across heterogeneous model variants of different sizes, transmitting compact semantic tokens rather than raw video data to reduce offloading delay. Furthermore, while recent LLM schedulers have begun utilizing textual prompts to drive dynamic routing decisions, this approach is fundamentally blind to the multimodal information inherent in video queries. Instead, QuickGrasp evaluates actual local execution results and leverages cross-modal semantic information to guide offloading configurations. This ensures that network and edge resources are strictly reserved for multimodal queries that require advanced reasoning.

\section{Conclusion}
\label{sec:conclusion}
The advent of VLMs has redefined the landscape of video querying services, transitioning systems from static pipelines toward interactive, open-world reasoning. Despite these capabilities, the severe resource requirements of modern VLMs have forced a compromise between the high response delay of remote-centric serving and the reduced accuracy of constrained local execution. This work introduces QuickGrasp, a responsive, QoS-aware system that bridges this gap through a local-first architecture with on-demand edge augmentation. To maximize system-wide efficiency, QuickGrasp integrates an accelerated video tokenizer to mitigate the video decoding bottlenecks, a confidence-based query router to trigger edge escalation, and a neural linear bandit to adaptively configure vision token density. Extensive prototype evaluation across diverse benchmarks confirms that QuickGrasp successfully delivers reasoning accuracy comparable to remote-hosted large VLMs while dramatically reducing response delays, establishing a highly practical and scalable architecture for next-generation video querying services.

\balance
\bibliographystyle{IEEEtran}
\bibliography{reference}

 




\end{document}